\documentclass{article}

\usepackage{microtype}
\usepackage{graphicx}
\usepackage{subcaption}
\usepackage{booktabs} 
\usepackage{multirow}
\usepackage{subcaption}

\usepackage{todonotes}
\setuptodonotes{fancyline, color=yellow!30}

\usepackage{hyperref}

\usepackage[accepted]{icml2026}

\usepackage{amsmath}
\usepackage{amssymb}
\usepackage{mathtools}
\usepackage{amsthm}
\usepackage{subcaption}

\usepackage[capitalize,noabbrev]{cleveref}

\theoremstyle{plain}

\theoremstyle{definition}

\theoremstyle{remark}

\icmltitlerunning{Pretraining EHR Foundation Models with Patient-Aware Sampling}

\begin{document}

\twocolumn[
  \icmltitle{
    Pretraining EHR Foundation Models with Patient-Aware Sampling
  }



  \icmlsetsymbol{equal}{*}

  \begin{icmlauthorlist}
    \icmlauthor{Joshua Placidi}{ic_computing,ic_bioeng,ai4health}
    \icmlauthor{Yuxuan ``Edison'' Liu}{ic_computing,ic_bioeng,ai4health}
    \icmlauthor{Jinpei Han}{ic_computing,ic_bioeng}
    \icmlauthor{Marek Rei}{ic_computing}
    \icmlauthor{A. Aldo Faisal}{ic_computing,ic_bioeng,ai4health,bayreuth}
  \end{icmlauthorlist}
  
  \icmlaffiliation{ic_computing}{Dept. of Computing, Imperial College London, London, United Kingdom}
  \icmlaffiliation{ic_bioeng}{Dept. of Bioengineering, Imperial College London, London, United Kingdom}
  \icmlaffiliation{ai4health}{UKRI Centres in AI for Health, United Kingdom}
  \icmlaffiliation{bayreuth}{Chair in Digital Health, Universit\"at  Bayreuth, Bayreuth, Germany}
  
  \icmlcorrespondingauthor{Joshua Placidi}{joshua.placidi25@imperial.ac.uk}
  \icmlcorrespondingauthor{A. Aldo Faisal}{aldo.faisal@imperial.ac.uk}

  \icmlkeywords{Machine Learning, Pretraining, Autoregressive, Health, Medical, ICML}

  \vskip 0.3in
]



\printAffiliationsAndNotice{}  

\begin{abstract}
Autoregressive foundation models for electronic health records (EHRs) typically inherit pretraining methods from language modeling, where patient trajectories are concatenated into a single token stream and windows are sampled from that stream.
In EHR data, this choice is consequential: windows may mix multiple patients, and patients with longer records contribute more optimization updates, potentially introducing bias.
We propose Patient Sampling, a pretraining sequence-construction method that allows us to control how training signal is distributed across patients. 
We compare this method to the standard approach, which we refer to as Global Stream.
We show that stochastic Patient Sampling with controllable weighting improves performance on real-world EHR data.
Across downstream clinical tasks on MIMIC-IV v2.2 and v3.1, Patient Sampling improves Macro AUROC and AUPRC over the Global Stream baseline.
These results identify training and validation sequence construction as important and underexplored design choices for autoregressive EHR foundation models.
\end{abstract}

\section{Introduction}

Autoregressive foundation models for electronic health records (EHRs) represent each patient history as a variable-length sequence of discrete tokens and train with a next-token prediction objective.
Recent work has shown that these models can support zero-shot clinical prediction by simulating future trajectories from observed patient histories \citep{renc2024ethos, foresight}.

\begin{figure}[t]
    \centering
    \includegraphics[width=\columnwidth]{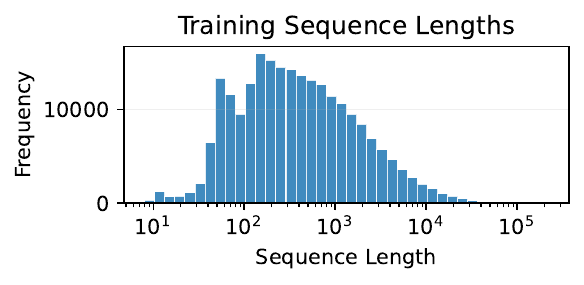}
    \caption{Distribution of patient token sequence lengths in the MIMIC-IV v2.2 training set, shown on a log-scaled x-axis.}
    \label{fig:train_sequence_length}
\end{figure}

Current EHR pretraining methods largely inherit sequence construction strategies from language modeling. In the standard setup, patient trajectories are concatenated into a global token stream and fixed-length training windows are sampled from that stream.
In EHR data, this is more than an efficiency choice: a window may contain events from multiple patients, and patients with longer records contribute more optimization updates during training.
This is especially relevant in our setting, where patient trajectory lengths are highly unequal, forming a log-normal distribution as shown in Figure~\ref{fig:train_sequence_length}.

\begin{figure}[t]
    \centering
    \includegraphics[width=\columnwidth]{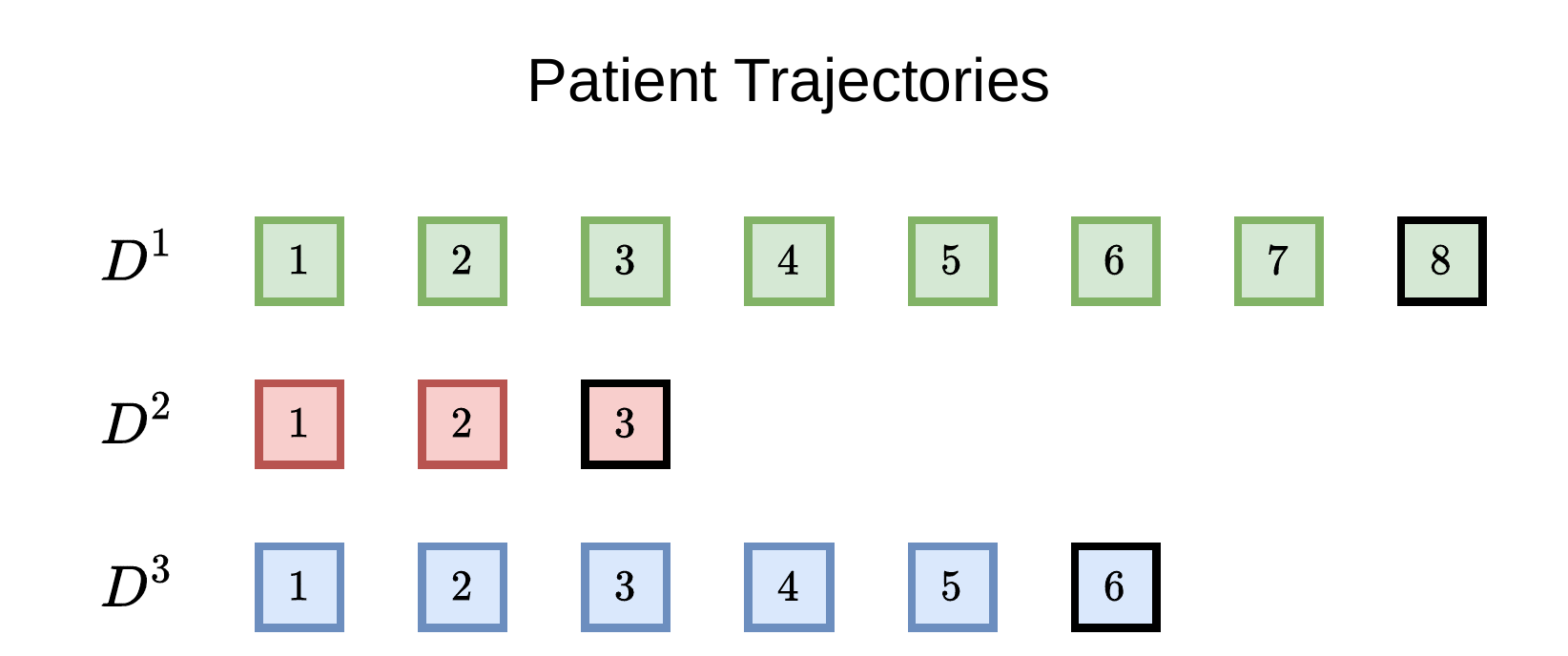}
    \caption{Toy example of an EHR dataset represented as variable-length token sequences $D_i$. Colored boxes denote EHR tokens, and black-bordered boxes indicate end-of-sequence tokens.}
    \label{fig:patient_trajectories}
\end{figure}

In this work, we study alternative methods for pretraining sequence construction. We identify three different approaches, which we refer to as Global Stream, Patient Chunks and Patient Sampling.

Keeping tokenization and model architecture fixed, we compare the standard Global Stream method with patient-aware alternatives that preserve patient boundaries and alter how training signal is distributed across patients. 
Across MIMIC-IV v2.2 and v3.1 \cite{PhysioNet-mimiciv-2.2, johnson2024mimiciv31, johnson2023mimicived22}, we find that Patient Sampling achieves stronger overall downstream performance than the Global Stream baseline, improving both Macro AUROC and Macro AUPRC.
Our findings highlight the impact of pretraining sequence construction methods, and to our knowledge this is the first work to explore this design choice in the context of autoregressive health foundation models.

\section{Background}

\begin{figure*}[!tb]
    \centering
    \includegraphics[width=\textwidth]{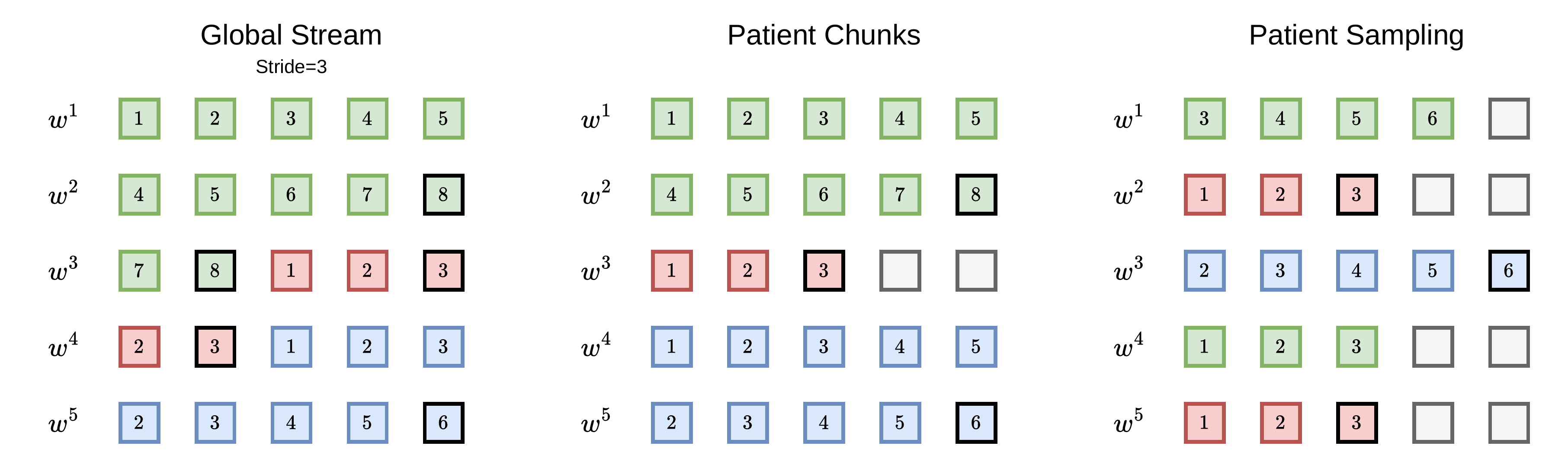}
    \caption{Comparison of the dataset constructions considered in this work. On the left 
        \emph{Global Stream} forms training windows from a global concatenated token stream. 
        \emph{Patient Chunks} constructs deterministic within-patient windows. 
        \emph{Patient Sampling} first samples a patient and then samples a window from that patient alone. 
        Colors denote different patients, and gray boxes indicate padding.}
    \label{fig:dataset_methods}
\end{figure*}

Autoregressive modeling has recently emerged as a promising paradigm for EHR data, where patient history is represented as a sequence of tokenized events.
Figure~\ref{fig:patient_trajectories} illustrates this setup with a toy tokenized EHR dataset.

\paragraph{EHR Foundation Models}
Recent work has applied autoregressive or generative pretraining to longitudinal EHR data. ETHOS \cite{renc2024ethos} and its follow-up ETHOS-ARES \cite{EthosARES} convert EHR data into tokenized patient timelines, train GPT-2 \cite{radford2019gpt2} style models with next-token prediction, and evaluate them by rolling out future events from a clinical prediction point to obtain zero-shot predictions for downstream tasks. Foresight \citep{foresight} models patient timelines with a generative pretrained transformer that combines structured EHR data and free text to predict future clinical concepts and outcomes. CoMET \cite{comet} demonstrates that autoregressive medical event pretraining with a next-token prediction objective scales to larger models and datasets with predictable scaling laws.
Together, these works establish autoregressive pretraining over patient timelines as a viable approach to EHR foundation modeling, but do not explicitly study pretraining sequence construction as a primary design variable.

\paragraph{Sequence Composition}
The impact of sequence composition has previously been explored in language model pretraining.
\citet{intra-document-masking-llms} show that concatenating unrelated documents into a fixed-length sequence can introduce distracting cross-document context, harming language modeling and downstream performance.
They demonstrate that both intra-document masking and retrieval-based packing of related documents improve performance.
Similarly, In-Context Pretraining \citep{shi2024incontext} shows that re-ordering documents so that each context window contains semantically related material improves tasks requiring stronger contextual reasoning across document boundaries.
These findings motivate studying sequence composition in EHR modelling, where patient trajectories are highly unequal in length and standard sequence construction may further amplify this imbalance.

\begin{table*}[t]
\centering
\small
\setlength{\tabcolsep}{3.5pt}
\resizebox{\textwidth}{!}{
\begin{tabular}{lll|cc|cc|cc|cc|cc}
\toprule
\multicolumn{3}{c|}{Dataset Method} 
& \multicolumn{4}{c|}{MIMIC-IV v2.2 + ED v2.2}
& \multicolumn{4}{c|}{MIMIC-IV v3.1 + ED v2.2}
& \multicolumn{2}{c}{} \\
\cmidrule(lr){1-3}
\cmidrule(lr){4-7}
\cmidrule(lr){8-11}
\cmidrule(lr){12-13}
Train & Validation & $\alpha$
& \multicolumn{2}{c|}{ICU Mortality}
& \multicolumn{2}{c|}{ICU Readmission}
& \multicolumn{2}{c|}{ICU Mortality}
& \multicolumn{2}{c|}{ICU Readmission}
& \multicolumn{2}{c}{Macro} \\
\cmidrule(lr){4-5} \cmidrule(lr){6-7} \cmidrule(lr){8-9} \cmidrule(lr){10-11} \cmidrule(lr){12-13}
& & & AUROC & AUPRC & AUROC & AUPRC & AUROC & AUPRC & AUROC & AUPRC & AUROC & AUPRC \\
\midrule
Global Stream & Global Stream & -- & 0.878 & 0.434 & 0.725 & 0.383 & 0.843 & 0.428 & 0.703 & 0.380 & 0.787 & 0.406 \\
Global Stream & Patient Chunks & -- & 0.866 & 0.428 & 0.717 & 0.389 & \textbf{0.852} & 0.439 & 0.712 & 0.382 & 0.786 & 0.410 \\
\specialrule{0.12em}{0.08em}{0.08em}
Patient Chunks & Patient Chunks & -- & 0.865 & 0.357 & 0.664 & 0.313 & 0.832 & 0.327 & 0.662 & 0.334 & 0.756 & 0.332 \\
\specialrule{0.12em}{0.08em}{0.08em}
\multirow{6}{*}{Patient Sampling} & \multirow{6}{*}{Patient Chunks} & 0   & \textbf{0.885} & 0.432 & 0.726 & 0.392 & 0.851 & 0.437 & 0.701 & 0.372 & 0.791 & 0.408 \\
 & & 0.2 & 0.848 & 0.350 & 0.721 & 0.382 & 0.839 & 0.402 & 0.690 & 0.365 & 0.774 & 0.375 \\
 & & 0.4 & 0.882 & 0.443 & 0.728 & 0.399 & 0.834 & 0.405 & 0.696 & 0.372 & 0.785 & 0.405 \\
 & & 0.6 & 0.884 & \textbf{0.456} & 0.730 & 0.405 & 0.850 & \textbf{0.446} & 0.706 & 0.381 & \textbf{0.792} & \textbf{0.422} \\
 & & 0.8 & 0.845 & 0.378 & 0.725 & 0.397 & 0.821 & 0.379 & 0.707 & 0.376 & 0.775 & 0.383 \\
 & & 1   & 0.863 & 0.416 & \textbf{0.733} & \textbf{0.407} & 0.848 & 0.435 & \textbf{0.719} & \textbf{0.399} & 0.791 & 0.414 \\
\bottomrule
\end{tabular}
}
\caption{Comparison of pretraining data constructions on ICU Mortality and ICU Readmission across MIMIC-IV v2.2 + ED v2.2 and MIMIC-IV v3.1 + ED v2.2. Results are reported as AUROC and AUPRC. The final macro column denotes the mean across all four reported task settings, and the best value in each metric column is shown in bold.}
\label{tab:dataset_results}
\end{table*}

\section{Method} \label{sec:method}

We consider a dataset of $N$ patients $\mathcal{D} = \{x^{(i)}\}_{i=1}^N$, where patient $i$ is represented by a sequence of EHR tokens
\[
x^{(i)} = \left(x^{(i)}_1, \dots, x^{(i)}_{L_i}\right),
\]
and $L_i$ is the sequence length of patient $i$. Following prior work on autoregressive EHR modeling, we train a decoder-only transformer with a next-token prediction objective on fixed-length token windows of length $S$.

We compare several strategies for constructing \emph{windows} $w = (w_1,\dots,w_S)$ for training and validation.
A window is a sequence of tokens that forms a single datapoint.

\paragraph{Global Stream.}
Our baseline follows the standard global stream construction used in autoregressive pretraining. We concatenate all patient sequences into a single stream
\[
z = x^{(1)} \mathbin{\|} x^{(2)} \mathbin{\|} \cdots \mathbin{\|} x^{(N)}
\]
with total length $T = \sum_{i=1}^N L_i$.
Given stride $r$, we define the set of valid start indices
\[
\mathcal{T}_{\mathrm{GS}} = \{1,\, 1+r,\, 1+2r,\, \dots,\, t_{\max}\},
\qquad t_{\max} \leq T-S+1.
\]
Each window is then
$w^{(t)} = (z_t,\dots,z_{t+S-1}),
\qquad t \in \mathcal{T}_{\mathrm{GS}}$.
This construction is simple and efficient, but a single window may contain tokens from multiple patients.

\paragraph{Patient Chunks.}
To preserve patient boundaries, we propose constructing windows independently within each patient sequence. For patient $i$, we define a deterministic set of chunk start indices
\[
\mathcal{B}_i = \{1,\, 1+S,\, 1+2S,\, \dots \}  \cap \{1,\dots,L_i\},
\]
We then add a final chunk that is right-aligned to the end of the sequence,
\[
b_{i,\mathrm{end}} = \max(1,\,L_i-S+1),
\]
and add this to the start index set
$
\mathcal{B}_i \leftarrow \mathcal{B}_i \cup \{b_{i,\mathrm{end}}\}
$. Each window is then
\[
w^{(i,b)} = \left(x^{(i)}_b,\dots,x^{(i)}_{\min(L_i,\,b+S-1)}\right),
\qquad b \in \mathcal{B}_i.
\]
Windows shorter than $S$ are right-padded to length $S$, and padded target positions are excluded from the loss and masked out. Unlike Global Stream, Patient Chunks always produces windows containing tokens from a single patient.

\paragraph{Patient-Aware Sampling.}
We next introduce a stochastic construction that decouples \emph{patient} and \emph{window} selection, to control the distribution of loss signal during training. 
This construction is only used for training due to its non-deterministic nature.

For each patient $i$, let $\mathcal{W}^{(i)}$ denote the set of valid start indices for constructing a window. We first sample a patient according to
\[
p_\alpha(i) = \frac{|\mathcal{W}^{(i)}|^\alpha}{\sum_{j=1}^N |\mathcal{W}^{(j)}|^\alpha},
\]
where $\alpha \in [0,1]$ controls the weighting of patients during training. When $\alpha=0$, patients are sampled uniformly; when $\alpha=1$, patients are sampled proportionally to the number of valid windows they contain, i.e.\ a patient with twice as many possible token windows as another patient will be sampled twice as often.

Given a sampled patient $i$, we then sample a start index uniformly from
\[
\mathcal{W}^{(i)} = \{1-S,\, 2-S,\, \dots,\, L_i-2\}.
\]
We clip the proposed start index to the first valid token $\tilde{s} = \max(1, s)$.
This choice of start indices allows tokens within a patient trajectory to appear with a broader range of left-context lengths, rather than systematically under-sampling early positions \footnote{Due to the need for an input and a target, the first and last tokens are still slightly under-sampled compared to interior tokens, though we believe this effect is minimal.}
.
Conditioned on $(i,\tilde{s})$, the resulting window is $x^{(i)}_{\tilde{s}:\tilde{s}+S-1}$, right-padded to length $S$ if necessary.

Global Stream may cross patient boundaries and introduce long-trajectory bias, Patient Chunks preserves patient boundaries but retains this imbalance, and Patient Sampling preserves boundaries while allowing the training distribution over patients to vary smoothly between patient-uniform and length-weighted sampling. Each method is visualized in Figure~\ref{fig:dataset_methods}.

\section{Experiments and Results}

\begin{table*}[t]
\centering
\small
\setlength{\tabcolsep}{4pt}
\begin{tabular}{l|cc|cc|cc|cc|cc}
\toprule
Method
& \multicolumn{2}{c|}{ICU Mortality}
& \multicolumn{2}{c|}{ICU Readmission}
& \multicolumn{2}{c|}{ICU Admission}
& \multicolumn{2}{c|}{Hospital Mortality}
& \multicolumn{2}{c}{Macro} \\
\cmidrule(lr){2-3} \cmidrule(lr){4-5} \cmidrule(lr){6-7} \cmidrule(lr){8-9} \cmidrule(lr){10-11}
& AUROC & AUPRC & AUROC & AUPRC & AUROC & AUPRC & AUROC & AUPRC & AUROC & AUPRC \\
\midrule
\multicolumn{11}{c}{MIMIC-IV v2.2 + ED v2.2} \\
\cmidrule(lr){1-11}
Global Stream
& 0.878 & 0.434 & 0.725 & 0.383 & \textbf{0.909} & \textbf{0.760} & 0.826 & 0.289 & 0.834 & 0.467 \\
Patient Sampling
& \textbf{0.884} & \textbf{0.456} & \textbf{0.730} & \textbf{0.405} & \textbf{0.909} & 0.756 & \textbf{0.846} & \textbf{0.297} & \textbf{0.842} & \textbf{0.478} \\
\midrule
\multicolumn{11}{c}{MIMIC-IV v3.1 + ED v2.2} \\
\cmidrule(lr){1-11}
Global Stream
& 0.843 & 0.428 & 0.703 & 0.380 & 0.905 & 0.748 & 0.813 & 0.308 & 0.816 & 0.466 \\
Patient Sampling
& \textbf{0.850} & \textbf{0.446} & \textbf{0.706} & \textbf{0.381} & \textbf{0.910} & \textbf{0.757} & \textbf{0.834} & \textbf{0.322} & \textbf{0.825} & \textbf{0.477} \\
\bottomrule
\end{tabular}
\caption{Comparison of Global Stream and Patient Sampling ($\alpha = 0.6$) across ICU Mortality, ICU Readmission, ICU Admission, and Hospital Mortality for MIMIC-IV v2.2 + ED v2.2 and MIMIC-IV v3.1 + ED v2.2. Results are reported as AUROC and AUPRC, and the Macro columns report the mean across the four tasks within each dataset version. Best value in each column is shown in bold.}
\label{tab:summary_results}
\end{table*}

\paragraph{Experimental setup.}
We evaluate the methods introduced in Section~\ref{sec:method} using two tokenized EHR datasets derived from MIMIC-IV: version 2.2 \cite{PhysioNet-mimiciv-2.2} and version 3.1 \cite{johnson2024mimiciv31}, both accessed via PhysioNet \cite{goldberger2000physionet}. In both settings, we also include the Emergency Department module, MIMIC-IV-ED version 2.2 \cite{johnson2023mimicived22}. Each dataset is split by patient into 80\% train, 10\% validation, and 10\% test sets.

We train a 6-layer GPT-2 model \cite{radford2019gpt2} on a single NVIDIA GH200 with sequence length $S=2048$.
Validation loss is evaluated every 10K training steps, and the checkpoint with the lowest validation loss is used for downstream evaluation.
Because Patient Chunks is deterministic and both patient-aware methods operate on single-patient windows, we use Patient Chunks as the validation dataset for Patient Chunks and Patient Sampling training datasets. For Global Stream, we report results with both Global Stream and Patient Chunks as validation datasets.
We evaluate downstream performance using the rollout framework of \citet{EthosARES} on the following clinical tasks: ICU Mortality, ICU Readmission, ICU Admission, and Hospital Mortality. See Appendix~\ref{app:framework} for full training and evaluation details.

\begin{figure}[t]
    \centering
    \includegraphics[width=\linewidth]{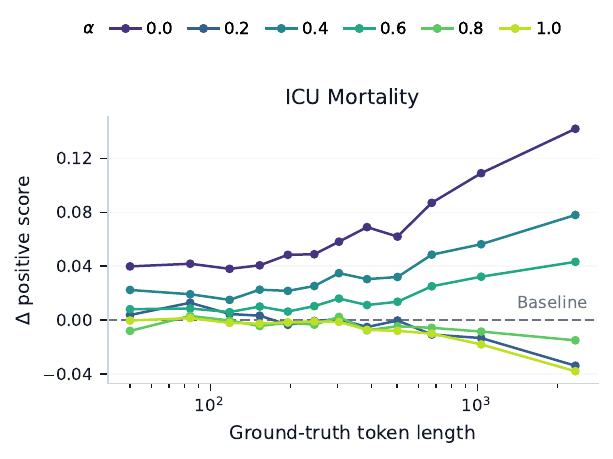}
    \caption{$\Delta$ positive score on ICU Mortality (MIMIC-IV v2.2) for Patient Sampling models with different $\alpha$ values, relative to a Global Stream baseline. Positive score is the fraction of simulations that terminate with the correct end token, delta positive score is measured relative to the Global Stream baseline. Datapoints are grouped by the number of tokens of the ground truth sequence.}
    \label{fig:alpha_ground_truth_length}
\end{figure}

\paragraph{Main results.}
We first evaluate each method on the ICU Mortality and ICU Readmission tasks, sweeping Patient Sampling over $\alpha$ values in increments of 0.2.
Results are shown in Table~\ref{tab:dataset_results}.
We find that Patient Chunks does not improve over the Global Stream baseline, suggesting that simply preserving patient boundaries is not sufficient.
Patient Sampling with $\alpha=0.6$ yields the strongest macro performance across both tasks in both datasets, and we therefore select this model for broader evaluation.

Table~\ref{tab:summary_results} shows results from further evaluation on the Hospital Mortality and ICU Admission tasks.
We find performance improvements on the Hospital Mortality task across both datasets. Improvements on ICU Admission are marginal, with no gain on MIMIC-IV v2.2 and small gains on MIMIC-IV v3.1.
Across all four benchmarks and both datasets, Patient Sampling outperforms Global Stream on all four macro metrics.

\paragraph{Effect of $\alpha$.}
Figure~\ref{fig:alpha_ground_truth_length} shows how the behavior of Patient Sampling varies with ground-truth token length on the ICU Mortality benchmark for MIMIC-IV v2.2. Smaller values of $\alpha$ produce the strongest positive shifts relative to the Global Stream baseline, with $\alpha=0$ showing the largest effect overall.
As hypothesized, $\alpha=0$ improves performance on shorter sequences, but this effect is actually amplified as ground-truth sequence length increases.
Intermediate settings such as $\alpha=0.4$ and $\alpha=0.6$ also remain positive across much of the range, whereas larger values of $\alpha$ stay closer to parity or become negative as trajectory length increases.
Notably, $\alpha=0.2$ underperforms relative to neighbouring values of $\alpha$, highlighting the need for further investigation into the effect of $\alpha$ on downstream tasks.

\section{Conclusion}

We studied how pretraining sequence construction affects autoregressive EHR foundation models, proposing a new Patient Sampling method.
We show that Patient Sampling improves over the Global Stream baseline across a broad set of downstream benchmarks.
These results suggest that patient boundaries alone are insufficient, but that patient-aware sampling and control over how training signal is distributed can improve autoregressive EHR pretraining.
More broadly, our findings identify sequence construction as an important design choice for autoregressive EHR foundation models.

\section*{Acknowledgements}
JP is supported by UK Research and Innovation (UKRI) AI Centre for Doctoral Training in Digital Healthcare grant number EP/Y030974/1, and by a philanthropic BUPA PhD Scholarship.
YL is supported by UK Research and Innovation (UKRI) Centre for Doctoral Training in AI for Healthcare (EP/S023283/1).
AAF is supported by the UK Research and Innovation (UKRI) Turing AI Fellowship grant (EP/V025449/1). MR \& AAF  acknowledge support of the UKRI AI programme, and the Engineering and Physical Sciences Research Council (EPSRC), for the AI Hub in Generative Models (EP/Y028805/1).
AAF \& MR acknowledge the use of resources provided by the Isambard-AI National AI Research Resource (AIRR). Isambard-AI is operated by the University of Bristol and is funded by the UK Government’s Department for Science, Innovation and Technology (DSIT) via UK Research and Innovation; and the Science and Technology Facilities Council (ST/AIRR/I-A-I/1023).
\clearpage

\bibliography{references}
\bibliographystyle{icml2026}

\newpage
\appendix
\onecolumn




\section{Model, Training, and Evaluation Framework}
\label{app:framework}

\paragraph{Model.}
We use a decoder-only transformer based on a GPT-2 style architecture. The model consists of a learned token embedding layer, a stack of transformer blocks with causal self-attention and MLP sublayers, a final layer normalization, and a linear language-modeling head. Rotary positional embeddings (RoPE) are used in place of learned absolute position embeddings. Models were trained on a single NVIDIA GH200 with a standard next-token prediction objective over tokenized EHR trajectories.
Table~\ref{tab:core_hparams} summarizes the core model and optimization hyperparameters used in our experiments.

\begin{table}[t]
\centering
\small
\setlength{\tabcolsep}{5pt}
\begin{tabular}{ll}
\toprule
\textbf{Component} & \textbf{Setting} \\
\midrule
Embedding dimension & 768 \\
Transformer layers & 6 \\
Attention heads & 12 \\
Context length & 2048 \\
Dropout & 0.1 \\
RoPE base & 10000 \\
Batch size & 64 \\
Optimizer & AdamW \\
Learning rate & $1.5 \times 10^{-4}$ \\
Weight decay & 0.1 \\
Warmup steps & 4000 \\
Hold steps & 136000 \\
Decay steps & 260000 \\
Final LR ratio & 0.05 \\
Validation interval & every 10K steps \\
\bottomrule
\end{tabular}
\caption{Core model and training hyperparameters used in our experiments.}
\label{tab:core_hparams}
\end{table}

\paragraph{Evaluation framework.}
Downstream evaluation is performed using a rollout-based benchmark framework. For each benchmark, evaluation prompts are constructed by identifying a benchmark-specific prediction point together with two terminating outcome tokens in each patient trajectory. The model receives the patient history up to the prediction point and is then rolled out autoregressively until a terminating token is generated or a maximum generation length of 4096 tokens is reached. For each prompt, we run 20 stochastic rollouts. A scalar score is computed as the fraction of rollouts that terminate with the different end tokens, and this score is used as the model output for binary classification. Simulations that exceed the 4096-token limit are terminated and excluded from benchmark calculations. AUROC and AUPRC are then computed from the resulting per-prompt scores. Table \ref{tab:benchmark_definitions} presents the definitions for each benchmark we report.

We note that the rollout-based evaluation is stochastic and can be noisy, and due to computational constraints we were unable to repeat all experiments across multiple random seeds.

\begin{table}[t]
\centering
\small
\setlength{\tabcolsep}{4pt}
\begin{tabular}{l|l|l}
\toprule
\textbf{Benchmark} & \textbf{Prediction point} & \textbf{Terminating tokens} \\
\midrule
ICU Mortality
& \texttt{ICU\_ADMISSION}
& \texttt{MEDS\_DEATH}, \texttt{ICU\_DISCHARGE} \\

ICU Readmission
& \texttt{ICU\_DISCHARGE}
& \texttt{ICU\_ADMISSION}, \texttt{HOSPITAL\_DISCHARGE} \\

ICU Admission
& \texttt{HOSPITAL\_ADMISSION}
& \texttt{ICU\_ADMISSION}, \texttt{HOSPITAL\_DISCHARGE} \\

Hospital Mortality
& \texttt{HOSPITAL\_ADMISSION}
& \texttt{MEDS\_DEATH}, \texttt{HOSPITAL\_DISCHARGE} \\
\bottomrule
\end{tabular}
\caption{Benchmark definitions used in the rollout-based evaluation framework.}
\label{tab:benchmark_definitions}
\end{table}

\end{document}